\documentclass[balance,upint,subscriptcorrection,varvw,mathalfa=cal=boondoxo,french,pdf-a,colorlinks]{asmeconf}

\hypersetup{%
	pdfauthor={Nicolas Gautier},
	pdftitle={Comparison of Robot Morphologies and Base Positioning for Welding Applications},                  
	pdfkeywords={Welding, architecture, evaluation},
	pdfsubject = {Describes the asmeconf LaTeX template},			  
	pdflicenseurl={https://ctan.org/pkg/asmeconf},
}
\usepackage{graphicx}
\usepackage{amsmath}
\usepackage{float}
\usepackage{physics}

\usepackage{multirow}

\begin{document}




\title{Comparison of Robot Morphologies and Base Positioning for Welding Applications}
 
%
%
%

\SetAuthors{%
	Nicolas Gautier \affil{1}\affil{2}, 
        Yves Guillermit \affil{1},
        Yazid Sebsadji \affil{1},
	Mathieu Porez \affil{2},
	Damien Chablat \affil{2}\CorrespondingAuthor{Damien.Chablat@cnrs.fr}       
	}

\SetAffiliation{1}{Weez-U Welding, 101 rue de Coulmiers, 44 000 Nantes, France }
\SetAffiliation{2}{Nantes Université, École Centrale Nantes, CNRS, LS2N, UMR 6004, 1 rue de la Noe, 44321 Nantes, France}


\maketitle





\begin{abstract}

This article undertakes a comprehensive examination of two distinct robot morphologies: the PUMA-type arm (Programmable Universal Machine for Assembly) and the UR-type robot (Universal Robots). The primary aim of this comparative analysis is to assess their respective performances within the specialized domain of welding, focusing on predefined industrial application scenarios. 
These scenarios encompass a range of geometrical components earmarked for welding, along with specified welding paths, spatial constraints, and welding methodologies reflective of real-world scenarios encountered by manual welders. The case studies presented in this research serve as illustrative examples of Weez-U Welding practices, providing insights into the practical implications of employing different robot morphologies. Moreover, this study distinguishes between various base positions for the robot, thereby aiding welders in selecting the optimal base placement aligned with their specific welding objectives. By offering such insights, this research facilitates the selection of the most suitable architecture for this particular range of trajectories, thus optimizing welding efficiency and effectiveness. A departure from conventional methodologies, this study goes beyond merely considering singularities and also delves into the analysis of collisions between the robot and its environment, contingent upon the robot's posture. This holistic approach offers a more nuanced understanding of the challenges and considerations inherent in deploying robotic welding systems, providing valuable insights for practitioners and researchers alike in the field of robotic welding technology.
\end{abstract}

\section{Introduction}
%
Robots have become integral components in various industries, offering versatile solutions to complex tasks ranging from manufacturing and logistics to healthcare and beyond \cite{Weiss_2021}. However, selecting the most suitable robot architecture remains a critical decision, influencing factors such as performance, flexibility, and adaptability to specific applications \cite{Celik1999Robotselection}.
In this article, we present a systematic process for comparing robot architectures in the context of welding tasks in making informed design choices such as those faced by the French company Weez-U Welding when designing its innovative teleoperated robotic solutions to optimize their performances to their custumer's applications. Currently, many performance indicators are used to characterize the performance of robots, such as the determinants, the condition number, or the dexterity \cite{angeles2003fundamentals}. However, even methods allowing the normalization of the Jacobian matrix with the characteristic length do not always reach consensus in the robotics community \cite{chablat2002kinetostatic}. Indeed, these criteria are often different from the criteria used by engineers to design a robot, for example, in selecting motors, gears, or drivers. However, a robot by its definition is not designed to perform a single trajectory \cite{siciliano2008springer}, as its main property is its ability to execute families of trajectories \cite{angeles2003fundamentals,KhalilDombre2002}. This article revisits new design criteria by integrating the welding workspace and considering collisions in the selection of robots for which we designed a specific simulation environment under ROS2 \cite{Quigley2009ROS}. 

After a brief introduction, we will provide an overview of the study's context for welding robotics and the company that will utilize the chosen robot. Next, the robot architectures along with their dimensions will be presented, including the welding torches. The case studies will then be introduced, outlining the positions of the robots within their respective environments. Finally, the results will be presented based on the case studies. A conclusion will close this article.

\section{Welding Process}\label{sec:welding_rules}
%
Since 2019, Weez-U Welding has offered teleoperation solutions that promote collaboration between welders and robots. These solutions combine the precision and reliability of a robot handling the torch (the welding tool) with the adaptability and intelligence of human control. By doing this, the company aims to eliminate the dangerous and arduous aspects of welding, while also strengthening the profession's attractiveness, which is facing recruitment difficulties  \cite{Ferraguti2023welding}.
\subsection{Welding Torch Positioning Parameters}
%
The welding process involves localized melting of the parts to be joined using an intense heat source \cite{Weman2011Welding}. One of the crucial steps to ensure a quality weld is to maintain an appropriate position of the welding torch along the entire trajectory. Incorrect positioning of the torch can compromise the quality of the weld bead: uneven heat distribution that can lead to cracks, undercuts, or pores in the weld, improper material deposition that can result in adhesion and weld strength issues, etc. Therefore, it is crucial to consider the robot's ability to maintain a precise and stable torch position to ensure optimal welding results. There are different values of torch positions and orientations depending on the applications, technologies used, type and thickness of the materials to be welded, etc.

\begin{figure}
    \centering
    \includegraphics[width=9cm]{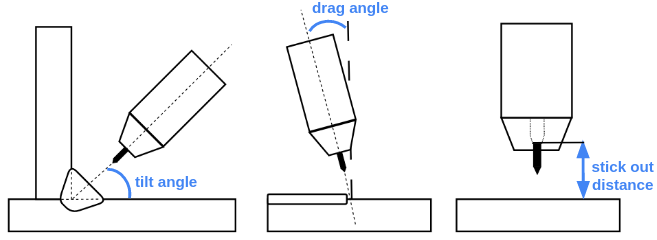} 
    \caption{Welding torch position and orientation parameters.}
    \label{fig:welding}
\end{figure}

\begin{itemize}
    \item The torch tilt angle (see Figure \ref{fig:welding}) which varies depending on the thickness of the parts but is usually between 40° and 50° for fillet weld (two perpendicular parts, see Figure \ref{fig:welding}). 
    \item The drag angle (see Figure \ref{fig:welding}) also varies depending on the processes and applications, usually ranging from -10° (pull) to +15° (push). 
    \item In MIG/MAG processes, a normal stick-out distance (distance between the contact tube outlet and the workpiece see Figure \ref{fig:welding}) is typically 10 to 15 times the electrode (wire) diameter, wire diameters are generally found between 0.6 mm and 1.6 mm. 
    
\end{itemize}

In this study, we will use the following set of generic parameters: 45° tilt angle (for plates with the same thickness), +7° drag angle (medium angle for aluminum welding ), and a 15 mm stick-out distance (medium distance for a 1.2mm diameter wire). The drag angle may not be maintained at the end of a corner, a transition phase may be performed over a maximum distance of 20 cm. These parameters were chosen based on the authors' previous experiences and feedback from their clients' applications.   \bigskip


Weez-U Welding's solution also focuses on real-time control of the welding process. The welder uses a remote controller to minimize exposure to toxic fumes, metal spatters, and radiation from the electric arc. A camera mounted on the torch allows the welder to see the weld in real-time. However, for optimal torch positioning and clear visibility of the weld pool, the camera needs to be aligned with the welding axis (ideally in front of the weld pool). This adds another constraint to the torch's orientation
%
\subsection{Welding Trajectories}
%
There are different types of welding trajectories, each adapted to specific needs in terms of shape, size, and weld quality \cite{Weman2011Welding}. The linear trajectory is one of the most common, used for straight welds horizontally or vertically. Circular trajectories are used for welding around cylindrical parts such as pipes. Weaving trajectories are often used for long welds and exhibit significant variations. Finally, other more complex trajectories may be used to address specific shapes that require a unique approach, such as a tubular connection formed by the intersection of two cylindrical parts. In terms of trajectory speed, a manual welder welds between 15 and 90 cm/min (2.5 mm/s to 15 mm/s).

In this study, we will only deal with linear and circular trajectories. All simulations will be carried out at the most critical speed of 90 cm/min. In cases where the limits of joint speed are exceeded, we will indicate the recommended maximum Cartesian speed (with a 20\% margin).
%
\subsection{Welding Workspace}
%
The targeted domains and markets of Weez-U Welding correspond to construction-type applications: i.e. transportation, construction, energy, etc. These workspaces are characterized by constrained environments where accessibility and mobility are limited. The robot considered in the following will be a lightweight arm that can be manually moved to the welding areas. Therefore, we will consider multiple base mounting positions for the robot. 
%
\section{Morphology of the Studied Robots}
%
\begin{figure}
    \centering
    \includegraphics[width=9cm]{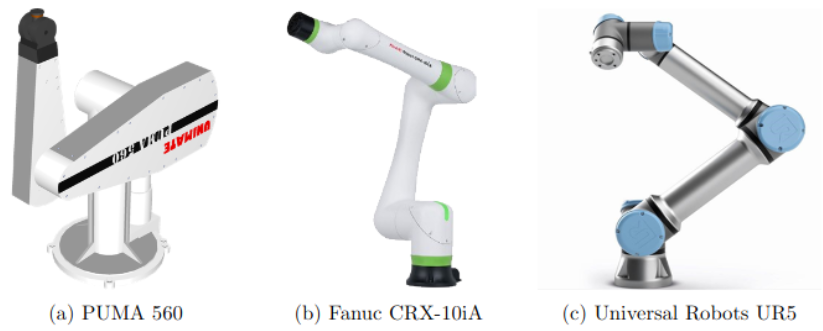} 
    \caption{Classic morphologies of 6-axis robotic arms.}
    \label{fig:morphologies}
\end{figure}
%
\subsection{Choice of Morphologies to Study}
\begin{figure}
    \centering
    \includegraphics[width=9cm]{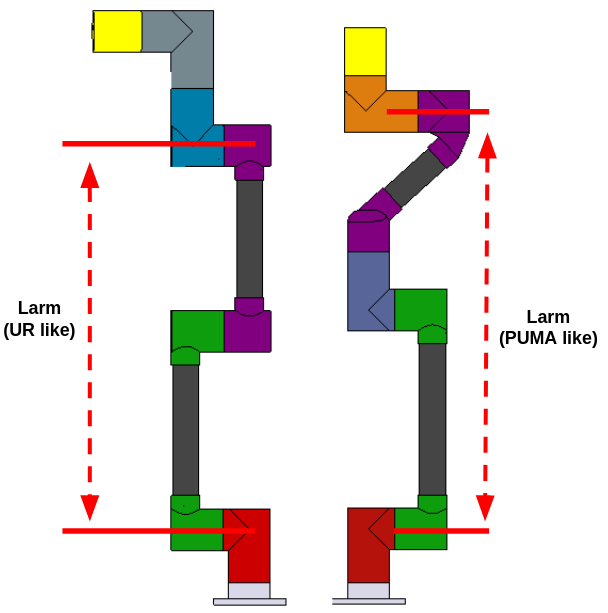} 
    \caption{3D models of the two studied morphologies.}
    \label{fig:3D}
\end{figure}

%
A study on around thirty commercially available 6-axis collaborative robots highlighted the predominance of three morphology types: anthropomorphic morphology (PUMA-like, Figure \ref{fig:morphologies}a), anthropomorphic with a wrist offset (Figure \ref{fig:morphologies}b), and with 3 parallel axes (UR-like, Figure \ref{fig:morphologies}c). The works of D. Salunkhe (see \cite{Salunkhe2023cuspidalrobot}) have shown that the second morphology presented constraints and peculiarities that made it less relevant for collaborative applications at the moment. Therefore, for this work, we decided to focus on comparing the other two morphologies. Figure~\ref{fig:3D} illustrates the 3D models of the two morphologies used in the simulations. Throughout the document, the notation $L_{\text{arm}}$ will refer to the lengths described in Figure~\ref{fig:3D}, corresponding respectively to 735 mm for the UR-like morphology and 793 mm for the PUMA-like arm. The segment lengths presented are the result of an internal study aimed at maximizing arm reach while remaining within the torque limits of our two types of actuator. The decision to offset the fifth actuator for the PUMA-like morphology is aimed at reducing wrist length to improve arm dexterity \cite{tourassis_comparative_1993}. The actuators considered in this study do not have articulation limits.
%
\subsection{Geometric Parameters of the two Studied Morphologies}
%
The Denavit Hartenberg (or DH) parameters of both studied morphologies are given in Table \ref{tab:DH_parameters} (see \cite{KhalilDombre2002} for the definition of frames for the anthropomorphic arm and \cite{hawkins_analytic_2013} for the definition of frames for the UR-like arm). \bigskip
\begin{table}[!ht]
\centering
\caption{DH Parameters for the studied Anthropomorphic arm and UR-like arm.}
\label{tab:DH_parameters}
\begin{tabular}{|c|c|c|c|c|c|c|}
\hline
& \multicolumn{3}{c|}{PUMA-like} & \multicolumn{3}{c|}{UR-like}\\
\hline
$i$ & $d_i$ [m] & $a_i$ [m] & $\alpha_i$ [rad] & $d_i$ [m] & $a_i$ [m] & $\alpha_i$ [rad]\\
\hline
0 & - & 0 & 0  & - & 0 & 0 \\
1 & 0.11 & 0 & $\pi/2$ & 0.117 & 0 & $\pi/2$ \\
2 & 0 & 0.42 & 0 & 0 & 0.38 & 0 \\
3 & 0.38 & 0 & $-\pi/2$ & 0 & 0.355 & 0 \\
4 & 0 & 0 & $\pi/2$ & -0.11 & 0 & $\pi/2$ \\
5 & 0 & 0 & $-\pi/2$ & 0.22 & 0 & $-\pi/2$ \\
6 & - & 0 & 0 & 0.19 & - & 0 \\
\hline
\end{tabular}
\end{table}

\begin{figure}
    \centering
    \begin{subfigure}[b]{0.4\textwidth}
        \centering
        \includegraphics[width=9cm]{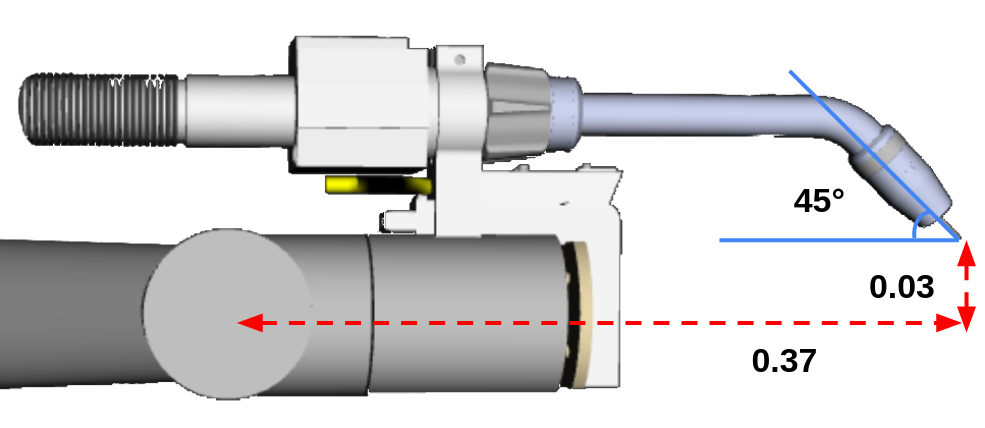} 
        \caption{PUMA-like arm.}
        \label{fig:Positioning1}
    \end{subfigure}
    \hfill
    \begin{subfigure}[b]{0.4\textwidth}
        \centering
        \includegraphics[width=9cm]{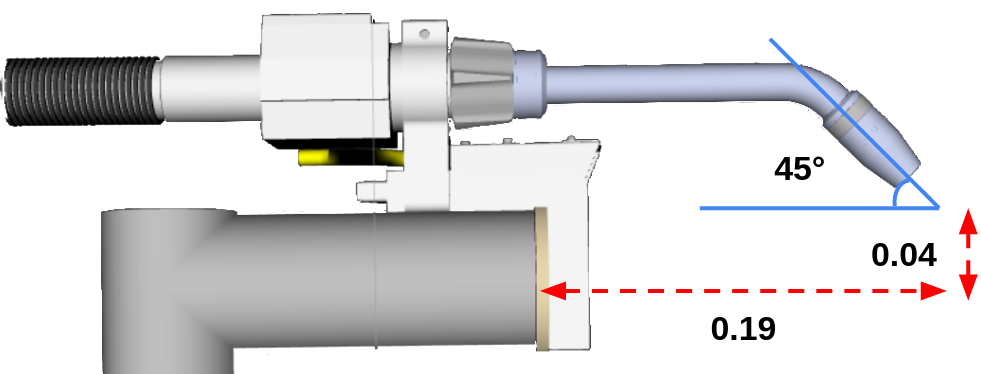} 
        \caption{UR-like arm.}
        \label{fig:Positioning2}
    \end{subfigure}
    \caption{Positioning of the welding torch on the PUMA-like arm and UR-like arm.}
    \label{fig:Positioning}
\end{figure}

The Tool Center Point (or TCP) of the PUMA-like arm is positioned at 0.37 m from the center of the wrist along the axis of the last actuator and with an offset of 0.03 m. The TCP of the UR-like arm is positioned at 0.19 m from the origin of the last actuator and with an offset of 0.04 m. The angle formed between the last axis and the stick out is 45° for both arms (see Figure \ref{fig:Positioning}).

%

\section{Application Cases}
%
In this section, we will describe a series of application cases representative of welding in the industry, highlighting challenges encountered during the assembly of various parts. The application cases addressed pertain to industrial applications currently performed by manual welders. For each presented use case, we will describe the methodology used by manual welders in real production conditions, as well as the specific criteria that will be used to evaluate the effectiveness of the robots. For the following figures from \ref{fig:case1} to \ref{fig:case6}, blue markers correspond to the different positions and orientations of the robot base fixation, and red curves correspond to the different weld beads on the workpiece. The base positions shown take into account the space constraints, the possible locations of base attachment points (beam or plate), the minimum distance between the TCP and the base, and the arm's reach limit. However, distances have not been optimized for every morphology and every trajectory but chosen in the way an operator is likely to do.  Moreover, for each played trajectory according to the time $t$, we define by $t_0$ and $t_n$ the initial and final times of the trajectory respectively. 
%
\subsection{Upward Trajectory with Base Position on the Ground}
\begin{figure}[!ht]
    \centering
    \includegraphics[width=8cm]{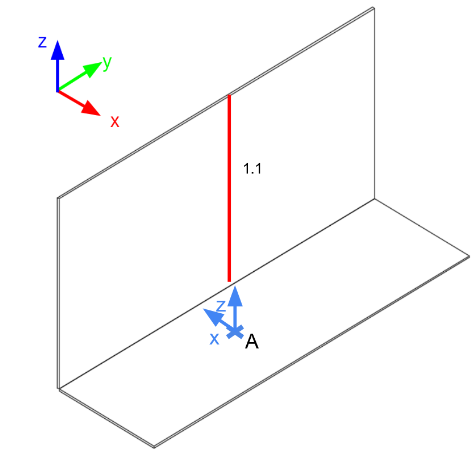} 
    \caption{Application Case 1: Upward trajectory with base position on the ground.}
    \label{fig:case1}
\end{figure}

%
In this application, the robot base is placed on the ground (see Table \ref{tab:robot_base_position}), and we aim to perform an upward trajectory from the ground (see Figure \ref{fig:case1}). Here, we seek to determine the maximum weldable height while maintaining the torch inclination constant. We aim to maximize the following criterion:
\begin{equation}
h = \frac{[z(t)]_{t_0}^{t_n}}{ L_{\text{arm}}}\text{ .}
\end{equation}
%

%
\subsection{Horizontal Trajectory with Base Position on Wall}
\begin{figure}[!ht]
    \centering
    \includegraphics[width=9cm]{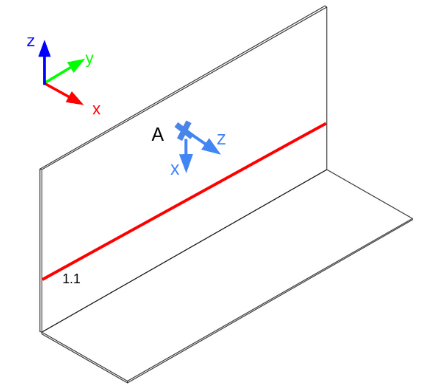} 
    \caption{Application Case 2: horizontal trajectory with the base position on the wall.}
    \label{fig:case2}
\end{figure}

%
In this application, the robot base is placed on a wall (see Table \ref{tab:robot_base_position}), and we aim to perform a horizontal trajectory (see Figure \ref{fig:case2}). Here, we aim to determine the maximum weldable length between singularity limits. We aim to maximize the following criterion:
\begin{equation}
d= \frac{[y(t)]_{t_0}^{t_n}}{2 L_{\text{arm}}}\text{ .}
\end{equation}
%
\subsection{Fillet weld around end}
\begin{figure}[!ht]
    \centering
    \includegraphics[width=9cm]{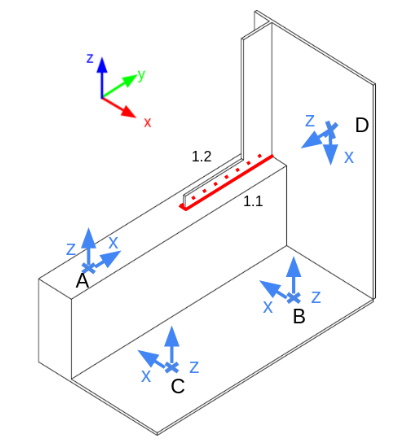} 
    \caption{Application Case 3: fillet weld around end.}
    \label{fig:case3}
\end{figure}

%
In this application, the workpiece is welded on both sides in a single trajectory (see Figure \ref{fig:case3}). Table \ref{tab:robot_base_position} summarizes the four studied base positions. The contouring point is the most critical step as it must be performed quickly to prevent melting of the metal piece (increase the Cartesian speed by 15\%). The change in torch orientation also requires significant articulation movements. We aim to minimize the joint speeds to avoid approaching the speed limits provided by the motors. The following criterion will be used:
\begin{equation}
v_{\text{max}}= v_{\text{cartesian}} | \max \left( \frac{\dot{q}_i}{\dot{q}_{i_{\text{max}}}}\right)=0.8, i \in \{1, 2, \ldots{}, 6\}\text{ ,}
\end{equation}
where $\dot{q}_i$ and $\dot{q}_{i_{\text{max}}}$ are the joint speed and the joint speed limit of the i$^{th}$ joint of the arm respectively.
%

%
\subsection{Pipe on plate welding Trajectory}
\begin{figure}[!ht]
    \centering
    \includegraphics[width=9cm]{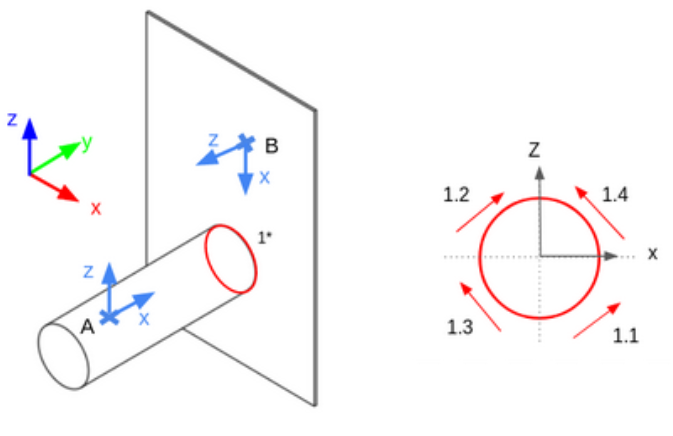} 
    \caption{Application Case 4: pipe on plate welding Trajectory.}
    \label{fig:case4}
\end{figure}

%
In this application, we aim to weld a tube with a diameter of 0.3 m (see Figure \ref{fig:case4}). The robot is either directly fixed on the tube or positioned above as a wall mount (see Table \ref{tab:robot_base_position}). The trajectory is performed in four steps (4/4 circle) by a manual operator, who welds in the following order 1.1, 1.2, 1.3, and 1.4 (see Figure \ref{fig:case4}). The application requires that all trajectories be performed without changing the base position. For symmetry reasons, the feasibility study can be summarized by evaluating trajectory 1.1 or 1.3 and trajectory 1.2 or 1.4. In this application, we will focus on the number of achievable trajectories with and without camera support.
%

%
\subsection{Beam T-connection with stiffener}
\begin{figure}[!ht]
    \centering
    \includegraphics[width=9cm]{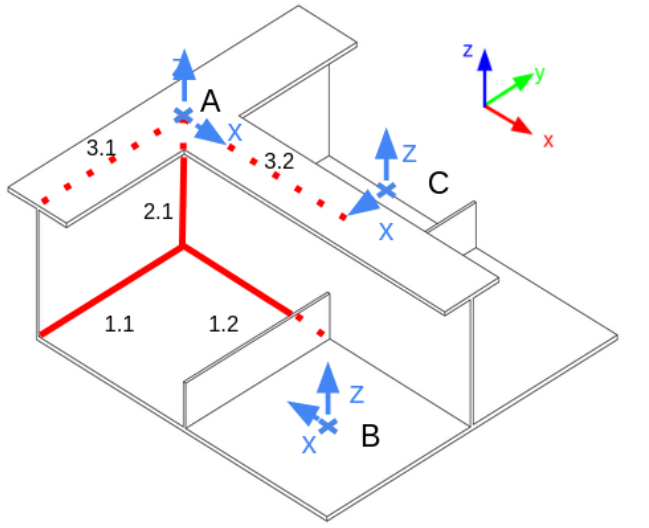} 
    \caption{Application Case 5: Beam T-connection with stiffener.}
    \label{fig:case6}
\end{figure}

%
This application features complex trajectories due to obstacles in the environment (see Figure \ref{fig:case6}). We will analyse achievable robot movements without relocating the base. If none of the pre-defined base positions (Table \ref{tab:robot_base_position}) allow all three trajectories, we will investigate if combining them can achieve the desired outcome.
\begin{table}[htbp]
\centering
\caption{Positioning of the Robot Base for the Application Cases.}
\label{tab:robot_base_position}
\begin{tabular}{|c|c|c|c|c|c|c|}
\hline 
Position & $X$  & $Y$ & $Z$  & Roll  & Pitch  & Yaw  \\
 & [m] &  [m] &  [m] &  [rad] &  [rad] & [rad] \\ \hline
\hline \multicolumn{7}{|c|}{\textbf{Application Case 1}}\\ \hline
A & 0.5 & -0.3 & 0 & 0 & 0 & $\pi$ \\
\hline \hline \multicolumn{7}{|c|}{\textbf{Application Case 2}}\\ \hline
A & 0 & 0 & 0.8 & 0 & $\pi/2$ & 0 \\
\hline \hline \multicolumn{7}{|c|}{\textbf{Application Case 3}}\\ \hline
A & 0 & -1 & 0.3 & 0 & 0 & $\pi/2$ \\
B & 0.3 & -0.3 & 0 & 0 & 0 & $\pi$ \\
C & 0.3 & -0.8 & 0 & 0 & 0 & $\pi$ \\
D & 0.3 & 0 & 0.5 & 0 & $\pi/2$ & $-\pi/2$ \\
\hline \hline \multicolumn{7}{|c|}{\textbf{Application Case 4}}\\ \hline
A & 0 & -0.5 & 0.15 & 0 & 0 & $\pi/2$ \\
B & 0 & 0 & 0.5 & 0 & $\pi/2$ & $-\pi/2$ \\
\hline \hline \multicolumn{7}{|c|}{\textbf{Application Case 5}}\\ \hline
A & 0 & 0 & 0.29 & 0 & 0 & 0 \\
B & 0.6 & -0.25 & 0 & 0 & 0 & $\pi$ \\
C & 0.2 & 0.25 & 0 & 0 & 0 & $-\pi/2$ \\
\hline
\end{tabular}
\end{table}
%

\section{Simulation Methodology}
%
\subsection{Comparison Criteria}
%
The methodology adopted to compare the two robot morphologies involves several essential criteria. First and foremost, it is crucial to ensure that the envisaged trajectory is feasible for the robot, this entails: 1) the absence of collisions with the environment and the absence of singularities; 2) that the torque required to execute the trajectory must remain within the limits of the robot's actuators; 3) as well as the joint velocities, which must remain below the motor limits. Once done, to evaluate the performance of each morphology, we will test each possible configuration and analyze the number of postures that allow the trajectory to be performed. The PUMA-type arm offers 8 solutions to the inverse kinematics problem, whereas the UR-type morphology offers either 2, 4, 6, or 8 solutions, depending on the case.  The results of the best configuration will be presented (maximizing specific application parameters and minimizing both the required motor torques and joint velocities). This comparison will be performed for all base positions of the robot defined for each application case. This will allow us to determine, for each morphology, the optimal placement of the base among the solutions.
%
\subsection{Simulation Methodology}
\begin{figure}[!ht]
    \centering
    \includegraphics[width=9cm]{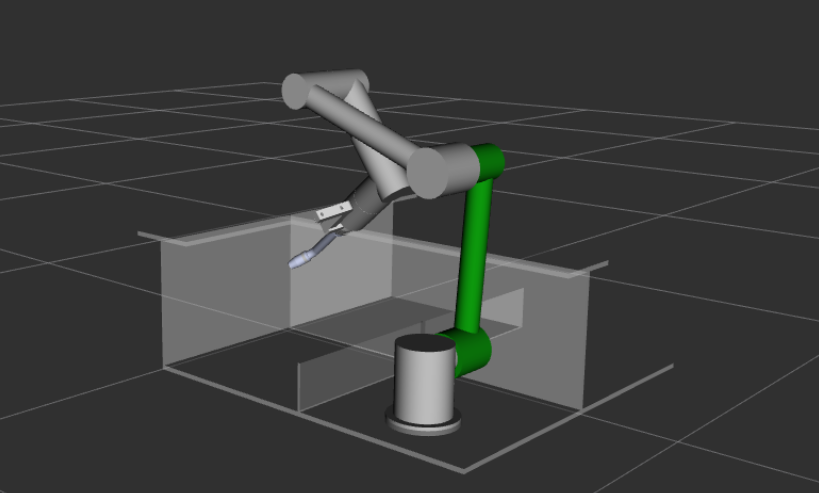} 
    \caption{Simulation visualization using Rviz.}
    \label{fig:rviz}
\end{figure}

\begin{figure*}[!ht]
    \centering
    \includegraphics[width=19.2cm]{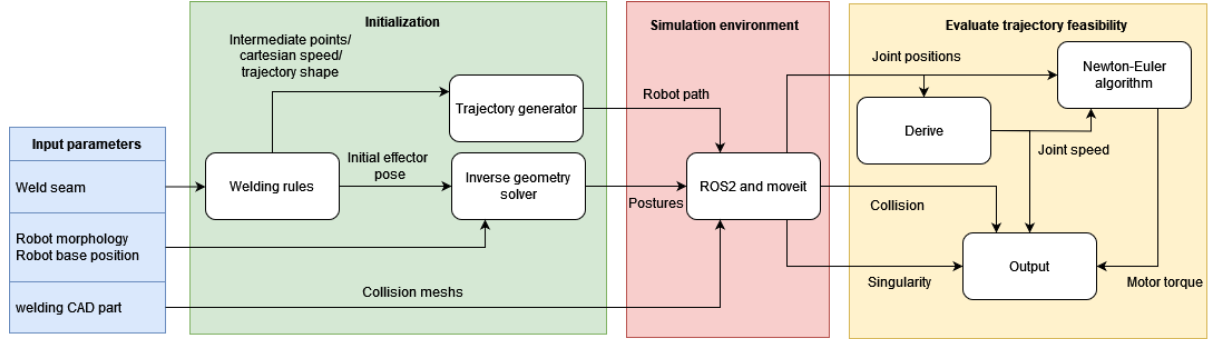} 
    \caption{Simulation methodology used to determine the feasibility of a trajectory.}
    \label{fig:ROS}
\end{figure*}

%
As mentioned before, the objective of the simulation is to determine the feasibility of a trajectory for a given robot morphology and base position. Our input data will therefore be (i) the weld bead to be performed, (ii) the robot morphology, (iii) its base position, and (iv) the CAD model of the piece to be welded. The welding rules presented in Section \ref{sec:welding_rules} allow us to construct the trajectory to be performed by the robot's end effector (welding torch). Knowing the initial position of the welding torch as well as the position of the robot's base, we can calculate all solutions to the inverse kinematics problem. Once done, all this information is transmitted to the ROS2 simulation environment \cite{Quigley2009ROS} where the generated trajectories are executed (see Figure~\ref{fig:rviz}). The different simulation variables are then processed to obtain the joint speeds and applied motor torques. Based on these two variables, as well as the presence of collision and singularity, we can establish the feasibility of the trajectory. Figure \ref{fig:ROS} summarizes the methodology used in this work.
%
\subsection{Simulation Tools}
%

\textbf{Inverse Kinematics Problem}: Both studied morphologies have analytical solutions to the inverse kinematics problem. The resolution methods can be found in \cite{KhalilDombre2002} for the PUMA-type arm and \cite{hawkins_analytic_2013} for the UR-type arm. \\

\textbf{Collision and Singularity Detection}: For this study, as mentioned before, we used a ROS2 simulation environment and the Moveit toolbox \cite{coleman2014reducing}. Moveit is used for motion control, as well as collision detection with the environment and collisions with the robot itself. We configured Moveit to detect collisions with a margin of 0.01 m. This ensures a minimum robustness regarding the placement of the base and differences in piece geometry. Moveit also informs when the robot enters singularity with a margin of 6°, beyond which the joint velocities become too high for the actuators. \\

\textbf{Joint Velocities}: During $t_0\leq t \leq t_n$, we retrieve the joint positions from a ROS topic. We derive these values over time to obtain the joint speeds. The speed limits considered for the motors are 5 rpm for the first three and 10 rpm for the last three. \\

\textbf{Motor Torques}: For the calculation of motor torques, we used the Newton-Euler method \cite{khalil_dynamic_2010}. We only considered static torques. Let's note that the speeds and accelerations are low during the welding process and actuator reduction ratios are very high (of the order of 100), the dynamic torques can be neglected.
For the calculation of the centers of mass, we assumed that the mass of each part is evenly distributed (except for the welding torch [0.03 m, 0, 0.08 m]). The mass of the welding hose package is considered as a 2 kg point mass applied to actuator 3. The center of mass applied to an actuator is calculated from the barycenter of the downstream parts. Table \ref{tab:masses-com} summarizes the new masses and positions of the centers of mass applied to each joint in their respective frame starting from the last actuator.

\begin{table*}[htbp]
\centering
\caption{Masses and center of mass positions used to calculate the motor torques.}
\begin{tabular}{|c|c|c|c|c|}
\hline
\textbf{Joint} & \textbf{UR-like Mass [g]} & \textbf{UR-like COM [m]} & \textbf{PUMA-like Mass [g]} & \textbf{PUMA-like COM [m]} \\ \hline
J1 & 1 040 & [0, -0.08, 0] & 1 040 & [0, 0.08, 0] \\ \hline
J2 & 3 190 & [0.362, 0, 0.05] & 3 210 & [0.4, 0, 0] \\ \hline
J3 & 890 & [0.295, 0, 0] & 1 040 & [0, 0.05, 0] \\ \hline
J4 & 740 & [0, 0, 0] & 860 & [0, 0.1, -0.05] \\ \hline
J5 & 740 & [0, 0, 0] & 740 & [0, 0, 0] \\ \hline
J6 & 2 000 & [0, -0.03, 0.08] & 2 000 & [-0.03, 0, 0.11] \\ \hline
\end{tabular}
\label{tab:masses-com}
\end{table*}
Finally, the maximum torque considered applicable to the actuators is 65 Nm for actuators 1, 2, and 3 and 20 Nm for actuators 4, 5, and 6.
%
\section{Results}
%
This section presents simulation results for various welding applications. Robot performance was assessed based on previously defined general and specific criteria. For each robot morphology and base position, we will focus on the best-performing posture, while also mentioning the number of postures achieving similar results. If achieving the desired welding parameters leads to infeasible trajectories, we will provide the closest achievable values (with a 20\% margin). Finally, we will identify the morphology and base placement that yielded the most successful outcomes.

In the following, the notations $\overline{\dot{q}}$, $\overline{\tau}$, $\dot{q}_{max}$ and $\tau_{max}$ refer respectively to the average motor speed, the average motor torque, the highest motor speed reached and the highest motor torque reached.
%
\subsection{Vertical trajectory with base position on the ground}
%
The main objective of this application is to determine the maximum weldable height starting from the ground while maintaining the torch inclination constant. The results obtained show that the UR-type arm reaches a maximum welding height of 720 mm ($\overline{\dot{q}}=1.96\%$, $\overline{\tau}=15.6\%$, $\dot{q}_{max}=9.8\%$ on motor 3 and $\tau_{max}=36\%$ on motor 4) for a single posture (depending on wrist position, performance for other postures was lower), while the PUMA-like arm reaches a maximum height of 560 mm ($\overline{\dot{q}}=3.34\%$, $\overline{\tau}=8.11\%$, $\dot{q}_{max}=51.93\%$ on motor 3 and $\tau_{max}=31.55\%$ on motor 2) for four postures. These values correspond respectively to $h=0.98$ and $h=0.71$ for criterion (1) defined in Section 4.1. The maximum heights were achieved by reaching a singularity position for both morphologies. These results demonstrate that the UR-like arm offers a superior vertical welding capability compared to the PUMA-like arm in this application. We can also note that the PUMA-type arm has higher joint speeds on this trajectory, with motor 3, in particular, reaching a maximum of 51.93\% of its limits.
%
\subsection{Horizontal trajectory with base position on the wall}
%
The main objective of this application is to determine the maximum horizontal distance weldable with the base position on the wall. The results obtained show that the UR-type arm reaches a maximum welding distance of 1500 mm ($\overline{\dot{q}}=3.43\%$, $\overline{\tau}=14.86\%$ and $\tau_{max}=68.66\%$ on motor 1) for two postures (depending on wrist position, performance for other postures was lower), while the PUMA-like arm reaches a maximum distance of 560 mm ($\overline{\dot{q}}=3.12\%$, $\overline{\tau}=12.53\%$ and $\tau_{max}=57.36\%$ on motor 1) for four postures. These values correspond respectively to $d=1.02$ and $d=0.76$ for criterion (2) defined in Section 4.2. The limit positions were achieved by reaching a singularity position for both morphologies. These results demonstrate that the UR-like arm offers a superior welding capability with positioning on the partition horizontally compared to the PUMA-like arm in this application. We can note that at the reach limit (when the arm is extended) the torque on motor 1 reaches high values for both morphologies (68.66\% for type UR and 57.36\% for type PUMA). However, this is still within the limits of the actuator.  
%
\subsection{Fillet weld around end}
%
The crucial step of this application is to perform the contouring of the workpiece without exceeding the motor's joint speed limits. The trajectories were made with a Cartesian speed of 90cm/min. The trajectory could not be performed in base positions A (self-collision between links 2 and 4, at the contouring point), B (collision between link 3 of the robot and the workpiece on trajectory 1.2), and D (collision between link 3 and the workpiece on trajectory 1.2 if the camera is positioned in front of the welding pool, or enter in singularity position if the camera is behind). In position C, the best posture for the UR-type morphology exceeds the maximum speed of motor 2 by 26.8\% ($v_{max} = 57cm/min$ for criterion (3), $\overline{\dot{q}}=7.11\%$, $\overline{\tau}=15.88\%$ and $\tau_{max}=39.84\%$ on motor 4). The best posture for the PUMA-like arm exceeds the maximum speed of motor 1 by 106.8\% ($v_{max} = 35 cm/min$ for criterion (3), $\overline{\dot{q}}=5.42\%$, $\overline{\tau}=9.93\%$ and $\tau_{max}=46.94\%$ on motor 2). 
%
\subsection{Pipe on plate welding trajectory}
%
In this application, we want to test the robots' ability to weld a 0.3 m diameter tube without changing the base position. The welding is divided into four trajectories. In position A, all trajectories can be performed with camera maintenance for the PUMA-like arm with only one posture for each trajectory ($\overline{\dot{q}}=7.65\%$, $\overline{\tau}=11.10\%$, $\dot{q}_{max}=22.27\%$ on motor 2 and $\tau_{max}=48.9\%$ on motor 2 for trajectory 1.1 \& 1.3 and $\overline{\dot{q}}=11.93\%$, $\overline{\tau}=8.7\%$, $\dot{q}_{max}=17.3\%$ on motor 1 and $\tau_{max}=42.02\%$ on motor 2 for trajectory 1.2 \& 1.4). For the UR-type arm, trajectories 1.2 \& 1.4 can be performed with camera maintenance for two postures ($\overline{\dot{q}}=11.07\%$, $\overline{\tau}=14.37\%$, $\dot{q}_{max}=27.39\%$ on motor 1 and $\tau_{max}=37.45\%$ on motor 4), but trajectories 1.1 \& 1.3 are not feasible even without maintaining the camera (collision with the tube). In position B, none of the trajectories are feasible for both robots. For trajectories 1.1 \& 1.3, the robots collide with the tube. For trajectories 1.2 \& 1.4, the robot enters in self-collision (link 2 and link 4). If the robot's base rises about 30 cm, trajectories 1.2 \& 1.4 could be feasible with camera maintenance.
%
\subsection{Beam T-connection with stiffener}
%
In this application, we want to test the robots' capability to weld in a highly constrained environment. For both morphologies, none of the trajectories can be realized with the chosen simulation parameters. However, trajectories 1 and 3 are achievable up to 75\% (up to halfway through the 2nd trajectory) with camera support. The remaining 25\% can be accomplished either by not maintaining the camera in line or by performing a second trajectory starting from the previous trajectory with the camera facing the weld pool (camera at the back on the first trajectory). Trajectory 2 is feasible if the drag angle is at least 12°; otherwise, the joint 6 axis collides with the T-structure. The camera will also become inaccessible at the end of the trajectory as the workpiece will obstruct the weld pool. For the UR-type arm, all trajectories can be performed from position A, as summarized in Table \ref{tab:trajectories}. The motor speeds and torques are far from the limits for all the trajectories with $\overline{\dot{q}}<5\%$, $\overline{\tau}<14\%$,  $\dot{q}_{max}<27\%$ and $\tau_{max}<37\%$ for all the motors. For the PUMA-type arm, the robot base will need to be moved to execute all trajectories (see Table \ref{tab:trajectories}), position A for trajectories 1 \& 3, and position B or C for trajectory 2 (no preferable solution). Similar to the UR-type arm, motors speeds and torques are far from the limits for all the trajectories with $\overline{\dot{q}}<6\%$, $\overline{\tau}<11\%$,  $\dot{q}_{max}<20\%$ and $\tau_{max}<54\%$ for all the motors.
\begin{table}[htbp]
    \centering
    \caption{Number of feasible trajectories for the UR-like and PUMA-like arms.}
    \begin{tabular}{|c|c|c|c|}
    \hline
    \multicolumn{4}{|c|}{UR-like} \\ \hline
    & Trajectory 1 & Trajectory 2 & Trajectory 3 \\ \hline
    Position A & 2 & 2 & 1 \\ \hline
    Position B & - & 1 & 1 \\ \hline
    Position C & - & 3 & - \\ \hline
    \end{tabular}
    \quad
    \begin{tabular}{|c|c|c|c|}
    \hline
    \multicolumn{4}{|c|}{PUMA-like} \\ \hline
    & Trajectory 1 & Trajectory 2 & Trajectory 3 \\ \hline
    Position A & 2 & - & 2 \\ \hline
    Position B & - & 4 & - \\ \hline
    Position C & - & 1 & - \\ \hline
    \end{tabular}
    \label{tab:trajectories}
\end{table}
%
\section{Conclusions}
%
This study compared the performance of two common industrial robot morphologies: the anthropomorphic arm (PUMA-type) and the three-joint parallel robot (UR-type). We evaluated their suitability for specific welding tasks using physical feasibility criteria like speed, motor torque, and collision avoidance. Simulated experiments analyzed the behavior of each morphology in various industry-relevant scenarios. The results revealed distinct advantages and limitations for each robot type, impacting their fit for different welding applications.

While the UR-type robot achieved superior performance in four out of five applications (as detailed in Table \ref{tab:comparaison}), its postures were less intuitive. Specific wrist configurations could hinder performance, as seen in applications 1 and 2 where optimal weld distances were achieved with only one or two postures. In contrast, the PUMA-type robot maintained consistent performance across various postures.

Interestingly, the study found no significant difference in average or maximum motor torques or joint speeds between the two morphologies (except for specific cases mentioned). However, the research did highlight the critical role of robot base positioning in optimizing welding processes. Recommendations for optimal base placement were provided for each simulated application.

Future work will explore the impact of welding tools (torch and hose) on robot dynamics. Additionally, an optimization algorithm will be developed to determine the ideal base placement for any given robot morphology and application within the feasible workspace. This will also involve evaluating the sensitivity of base placement to assess robot robustness.

\begin{table}
    \centering
    \caption{Relative performance of the UR-like arm compared with the PUMA-like arm based on the application cases criteria}
    \begin{tabular}{|c|c|c|c|c|c|}
    \hline
        Application cases & 1 & 2 & 3 & 4 & 5\\ \hline
        Relative performance [\%] & +38 & +34 & +63 & -50 & +50\\ \hline
    \end{tabular}
    \label{tab:comparaison}
\end{table}

\section*{Acknowledgement}
This research was supported by ANRT CIFRE grant n°2023 /1565 which funded the first author's doctoral studies. 
\bibliographystyle{asmeconf}  
\bibliography{asmeconf-sample}
\end{document}